\documentclass[sigconf, nonacm]{acmart}

\usepackage{listings}
\newcommand\vldbyear{2026}
\newcommand\vldbworkshop{Agent+Graph}
\newcommand\vldbauthors{\authors}
\newcommand\vldbtitle{\shorttitle} 
\newcommand\vldbavailabilityurl{https://github.com/zach-blumenfeld/aip-skillbench}
\newcommand\vldbpagestyle{empty}

\begin{document}
\title{AIP: A Graph Representation for Learning and Governing Agent Skills}

\author{Zach Blumenfeld}
\affiliation{%
  \institution{Neo4j}
  \state{MD}
  \country{USA}
}
\email{zach.blumenfeld@neo4j.com}

\author{Jim Webber}
\affiliation{%
  \institution{Neo4j}
  \city{London}
  \country{UK}
}
\email{jim.webber@neo4j.com}

\begin{abstract}
Agent Skills today consist largely of free-form prose requiring the agent to read, interpret, and re-derive how to act in every session. This imposes two compounding costs: reduced reliability on implementation-heavy tasks, and difficulty in skill creation and improvement. Editing prose is a fragile process that both humans and agents struggle with, particularly for domain-specific procedural knowledge that is underrepresented in model training. The Agent Instruction Protocol (AIP) addresses both by modeling a skill as a directed execution graph with discrete steps as nodes backed by deterministic scripts or natural-language descriptions, connected by explicit typed input/output edges, all governed by a schema-validated YAML specification. A compiler meta-skill translates existing human-written skills into this form.

The benefits are twofold. Firstly, compiling human-written skills to AIP raised Claude Sonnet's mean task reward from 0.60 to 0.71 and pass rate from 53\% to 67\% across 27 real agent tasks from SkillsBench---a statistically significant gain (Wilcoxon signed-rank $p = 0.011$), winning 12 tasks to 2 with 13 ties. It is often faster since the graph delivers vetted, runnable units to the agent rather than asking it to re-derive code, commands, and tool calls from natural language. Secondly, on creation and improvement, because each skill is schema-validated, functionally testable, and addressable node-by-node, failures can be diagnosed and repaired precisely. Two authored-skill failures were traced to the script level. After adjusting the AIP spec and recompiling, both recovered with zero regressions (one task going from 0/5 to 5/5), turning skill improvement into a measurable tuning loop rather than a prose rewrite. That same graph structure supports corpus-level governance and skill introspection, and provides a natural action space for reinforcement learning over skills.
\end{abstract}

\maketitle

\pagestyle{\vldbpagestyle}
\begingroup\small\noindent\raggedright\textbf{VLDB Workshop Reference Format:}\\
\vldbauthors. \vldbtitle. VLDB \vldbyear\ Workshop: \vldbworkshop.\\ 
\endgroup
\begingroup
\renewcommand\thefootnote{}\footnote{\noindent
This work is licensed under the Creative Commons BY-NC-ND 4.0 International License. Visit \url{https://creativecommons.org/licenses/by-nc-nd/4.0/} to view a copy of this license. For any use beyond those covered by this license, obtain permission by emailing \href{mailto:info@vldb.org}{info@vldb.org}. Copyright is held by the owner/author(s). Publication rights licensed to the VLDB Endowment. \\
\raggedright Proceedings of the VLDB Endowment. 
ISSN 2150-8097. \\
}\addtocounter{footnote}{-1}\endgroup

\ifdefempty{\vldbavailabilityurl}{}{
\vspace{.3cm}
\begingroup\small\noindent\raggedright\textbf{VLDB Workshop Artifact Availability:}\\
The source code, data, and/or other artifacts have been made available at \url{\vldbavailabilityurl}.
\endgroup
}

\section{Introduction}
\label{sec:intro}

\emph{Agent Skills}~\cite{anthropic-agent-skills}, introduced by Anthropic in 2025 and
subsequently released as an open standard, have become the dominant representation
for packaging reusable agent capabilities. A skill is a directory built around a
\texttt{SKILL.md} file: YAML frontmatter declaring a \texttt{name} and \texttt{description},
followed by natural-language instructions, optional scripts, and reference files that the
agent loads on demand through \emph{progressive disclosure}. The appeal of the format is
that it lets a broad range of domain experts transfer
procedural knowledge to AI agents in a lightweight, human-readable form~\cite{bakal2026knowledge}. Moreover, curated skills measurably raise task success across diverse domains~\cite{skillsbench}.

Yet a \texttt{SKILL.md} is a (mostly) free-form markdown document, and the skill as a
whole is consumed as natural-language content. As a result, the representation
inherits several limitations that become acute as agents are deployed on harder,
implementation-heavy tasks:

\begin{enumerate}
  \item \textbf{Skills neither capture nor enforce structure where warranted, leaving speed and reliability on the table.} Some context and judgment resist formalization, but a large share of a skill's procedural knowledge could be expressed
  as runnable code and explicit graphs of workflow steps, which is increasingly the way
  reliable agents are built~\cite{langgraph,google-adk,anthropic-effective-agents}. Skills allow this logic to remain in free prose for the agent to derive at runtime, so on
  every new session an agent re-plans the code, commands, and tool calls the task needs. On
  implementation-heavy tasks this is slow and token-intensive, and lets the agent take
  different, sometimes erroneous, paths that reduce reliability and increase run-to-run variance.
  Offloading deterministic steps to vetted, runnable
  code~\cite{gao2023pal,chen2023pot} and explicitly structured
  workflows~\cite{anthropic-effective-agents} instead is known to improve
  reliability. What is missing is a mechanism that reliably compiles this structurable
  knowledge, including the deterministic steps and the data that flows between them, into vetted,
  runnable code and an explicit step graph, so the agent need not re-derive it on every run.

  \item \textbf{Skill creation and tuning remains a slow, recurring process.} A skill must be tuned like a prompt. A skill consisting of mostly free-form prose is read and re-interpreted by the agent on every run, so it inherits the well-documented brittleness of prompts: small, semantics-preserving changes in wording or formatting can swing task
  accuracy by tens of points~\cite{sclar2024formatspread} and so it is difficult and error-prone to derive a skill that consistently works correctly.

  \item \textbf{Agent-assisted and self-improvement of skills remains a challenge.} A skill is only worth shipping when it encodes procedural knowledge lacking in model training~\cite{skillsbench,bakal2026knowledge}. Experts therefore author skills, but, because of the challenges presented above, increasingly enlist agents to help \emph{revise} them. The field as a whole is pushing toward agent self-improvement and reinforcement learning over skills~\cite{gao2025selfevolving,zweiger2025seal,robeyns2025selfimproving,xu2026agentskills}. This is hard for two compounding reasons: the material is unfamiliar by construction---the agent is revising domain-specific procedures it does not itself know---and free-form prose gives it no bounded surface to edit against. With nothing to constrain the edit, the agent's additive bias runs unchecked: outputs skew toward length and verbosity~\cite{singhal2024length,zhang2024verbosity}, and agents over-engineer and accrete rather than refine~\cite{licorish2025comparing}. Skills balloon and grow convoluted, intent drifts, and a fix in one place silently breaks another, while intrinsic self-revision without external feedback often leaves quality unchanged or degraded~\cite{huang2024selfcorrect,xu2024selfbias}. Without a more structured representation, skill improvement has neither a strong natural feedback loop nor a bounded action space.

  \item \textbf{Cause, effect, and governance are hard to establish.} Free-form prose has no
  clear addressable units. When a skill produces a wrong result there is no node, step, or typed
  value to which the failure can be precisely attributed, and a corpus of prose skills cannot be
  systematically audited for missing validation or approval steps. This is contrary to the
  traceability and accountability that the governance of agentic systems at scale
  demands~\cite{saini2026governing}.

\end{enumerate}

The Agent Instruction Protocol (AIP)\footnote{Today AIP is realized as a \emph{specification}, a schema-validated execution-graph format that an agent reads into context and follows. We retain the term \emph{protocol} for the typed contract this format defines, and for a runtime that would enforce graph traversal and local and remote skill calls. We set that out as future work (Section~\ref{subsec:protocol}).} addresses these issues by modeling a skill as a
directed execution graph. Discrete steps become nodes, each backed by a deterministic
script or, where human-like judgment is required, a natural-language description. Nodes are
connected by typed input/output edges, and the whole structure is governed by a
schema-validated YAML specification. A compiler meta-skill allows agents to translate
existing human-written skills into this form at authoring time, surfacing ambiguities, type
errors, and field inconsistencies before they can cause runtime failures. Crucially, AIP
does not displace human authorship. Experts still write the skills, and the compiler
translates their human-written source into a typed, script-backed surface that improves
how reliably agents \emph{execute} skills today. We also argue this is a better substrate for agent-assisted improvement and reinforcement learning in the future.

We evaluate AIP against human-curated skills on SkillsBench~\cite{skillsbench}, a benchmark
of 94 agent tasks across 8 domains, using a stratified 27-task sample with Claude Sonnet
as the solver. Compiling skills to AIP produced a statistically significant improvement in
task reward (Wilcoxon signed-rank $p = 0.011$), and we found that failures in AIP skills can
be diagnosed and corrected at the node level, given coded scripts connected by a clear,
typed execution-graph I/O structure, without causing regressions elsewhere. This suggests that the
benefits of AIP may compound for agent self-improvement and reinforcement learning (RL),
where the typed execution graph provides a bounded, validity-gated action space for learning
over skills~\cite{sutton2018rl}.

\textbf{Contributions.} Our work makes the following contributions:
\begin{itemize}
  \item \textbf{A multi-mode benchmark harness.} An extension to SkillsBench for comparing
  skill \emph{formats} and skill \emph{authoring methods} under a real solver, enabling
  controlled head-to-head evaluation.

  \item \textbf{Empirical evidence of improved task reward.} Compiling human-written skills
  to AIP significantly improves task reward across 27 tasks (mean reward
  $0.599 \rightarrow 0.705$, $+0.106$; Wilcoxon $p = 0.011$), with the mechanism localized to
  executability and procedural consistency.

  \item \textbf{A proof-of-concept for node-level diagnosis and repair.} During early experimentation, two compiler-authored
  failure modes were diagnosed at the node and script level by an agent (Claude Code). In this case they were
  corrected by editing the AIP specification, and recovered with zero regressions, but the tooling demonstrates strong potential for
  a diagnosis--edit--recompile--evaluate loop with a measurable reward signal---a working substrate for agent-assisted improvement and, prospectively,
  autonomous self-improvement and reinforcement learning over skills.

  \item \textbf{A path to corpus-level governance and inspection.} Because each skill is a
  typed, schema-validated graph, a library of AIP skills is queryable and inspectable: missing
  validation or approval steps can be audited, shared sub-procedures discovered, and skills
  composed from reusable node templates. Projected into a graph database, the same structure
  supports access control and visual introspection---a skill's steps and their typed
  input/output rendered as a graph (Figure~\ref{fig:aipgraph})---moving governance from manual
  documentation review to a structured query over a typed graph. As agents grow more
  autonomous and begin running their own self-improvement loops, this inspectable surface is
  what keeps human oversight and understanding tractable.
\end{itemize}

\section{Background}

\begin{figure}[t]
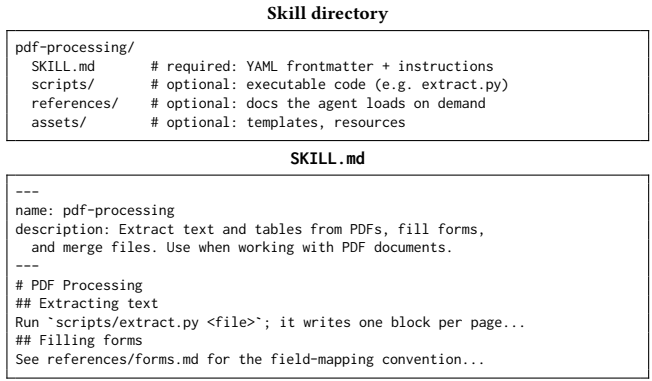

\centering
{\footnotesize\textbf{Skill directory}}\\[2pt]
\begin{lstlisting}
pdf-processing/
  SKILL.md       # required: YAML frontmatter + instructions
  scripts/       # optional: executable code (e.g. extract.py)
  references/    # optional: docs the agent loads on demand
  assets/        # optional: templates, resources
\end{lstlisting}
{\footnotesize\textbf{\texttt{SKILL.md}}}\\[2pt]
\begin{lstlisting}
---
name: pdf-processing
description: Extract text and tables from PDFs, fill forms,
  and merge files. Use when working with PDF documents.
---
# PDF Processing
## Extracting text
Run `scripts/extract.py <file>`; it writes one block per page...
## Filling forms
See references/forms.md for the field-mapping convention...
\end{lstlisting}
\caption{An Agent Skill is a directory built around a \texttt{SKILL.md} file---YAML frontmatter (required \texttt{name} and \texttt{description}) followed by free-form Markdown instructions---alongside optional \texttt{scripts/}, \texttt{references/}, and \texttt{assets/} that the agent loads only as a task requires (\emph{progressive disclosure})~\cite{agentskills-spec}.}
\label{fig:agentskill}
\end{figure}

Modern LLM agents carry out tasks by interleaving natural-language reasoning with calls to external tools---running code, querying systems, editing files---and observing the results~\cite{yao2023react,schick2023toolformer}. Because many tasks recur, these agents increasingly draw on reusable \emph{skills}: packaged procedural knowledge---instructions, and sometimes code---that tells them how to carry out a class of tasks~\cite{wang2023voyager}. Anthropic's Agent Skills standardize this as a \texttt{SKILL.md} file in a directory of optional resources the agent loads on demand (Figure~\ref{fig:agentskill})~\cite{anthropic-agent-skills,agentskills-spec}.

A skill's procedure lives largely in the free-form Markdown body of its \texttt{SKILL.md} (Figure~\ref{fig:agentskill}), which the agent reads and interprets at runtime. Though the skill format permits bundled scripts, in practice much of the procedure stays in prose.

\textbf{Graphs and structured agent workflows.} Agent development frameworks already represent agent workflows as graphs whose nodes are discrete steps and whose edges carry the data passed between them. This includes frameworks such as LangGraph and Google's Agent Development Kit (ADK)~\cite{langgraph,google-adk}. Earlier than that, DSPy, a Python framework for building AI systems, used declarative pipelines that can be compiled rather than hand-written to automatically optimize how language models are prompted (or fine-tuned), replacing brittle, manually tuned prompt templates~\cite{khattab2023dspy}. Anthropic, the AI safety and research company behind the Claude family of models and the agent skill spec, also recommends predefined, structured workflows where predictability and reliability matter~\cite{anthropic-effective-agents}. Dedicated benchmarks likewise find that supplying agents with explicit workflow structure improves their planning~\cite{flowbench}. Evidence that this matters comes from the same benchmark we use, SkillsBench, where the skills that help most pair focused guidance with executable code and reference files, while sprawling, comprehensive prose can actually hurt performance~\cite{skillsbench}. It is structure, not just content, that shapes how well a skill works. AIP brings this graph view to the skill itself, where a skill becomes an execution graph of typed steps, each backed by a script or a natural-language description.

\textbf{Measuring skills.} We build on SkillsBench, a containerized benchmark that isolates the effect of a skill. Each task pairs a natural-language instruction with a sandboxed environment and a programmatic verifier, run through the BenchFlow SDK~\cite{benchflow}, and is scored under three conditions: no skill, a human-curated skill, and a skill the agent generates for itself. Measuring task success across these conditions lets a skill's contribution be quantified directly, which our experiments extend with AIP.

\section{The Agent Instruction Protocol (AIP)}
\label{sec:aip}

\begin{figure*}[t]
  \centering
  \includegraphics[width=0.92\textwidth]{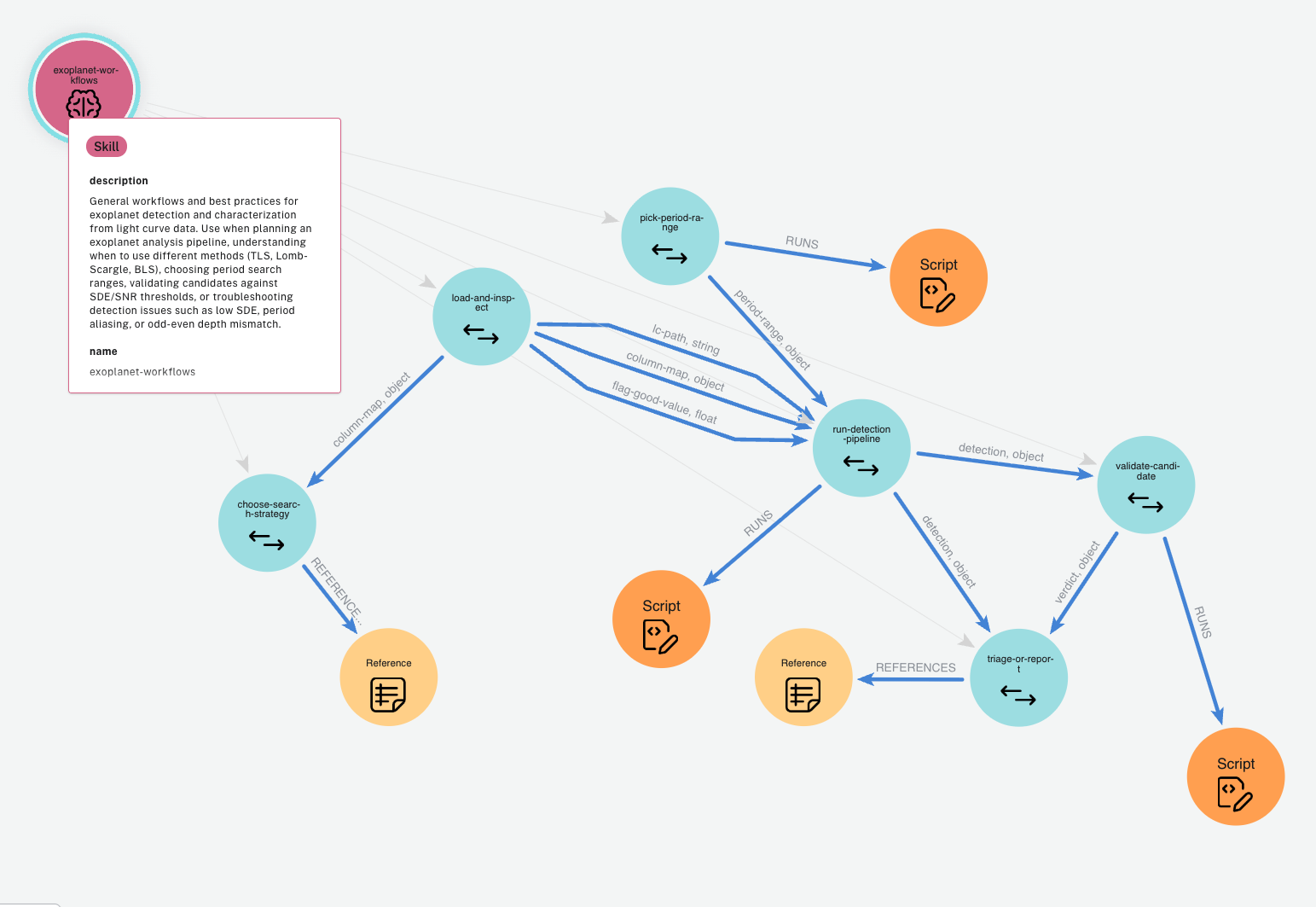}
  \caption{A compiled AIP skill (\texttt{exoplanet-workflows}) as a directed execution graph, projected into Neo4j. The pink \textsc{Skill} node carries the name and description; teal nodes are procedure \emph{steps}; arrows between steps are \emph{typed input/output edges} (e.g.\ \texttt{lc-path,\,string}; \texttt{detection,\,object}) that make data flow explicit and checkable. A \textsc{Runs} edge binds a step to a deterministic \emph{script} (orange); a \textsc{References} edge attaches a prose \emph{reference} (yellow) for steps needing judgment. The same typed structure is what makes a skill node-addressable, and queryable (Section~\ref{subsec:governance}).}
  \label{fig:aipgraph}
\end{figure*}

AIP is an extension of the Agent Skills specification~\cite{agentskills-spec}, one that defines a skill as a directed execution graph---today a specification, with the enforcing protocol left to future work (Section~\ref{subsec:protocol}). Nodes represent discrete operational steps where each step is either backed by a deterministic script for computational work, or described in natural language for steps requiring judgment or interaction. Step nodes are connected by typed input/output edges that make data flow explicit and checkable, and the entire structure is governed by a schema-validated YAML specification. Figure~\ref{fig:aipgraph} shows a real compiled skill rendered as a graph. The major node and edge types are as follows:

\vspace{\baselineskip}  
\noindent\begin{minipage}{\columnwidth}
\begin{lstlisting}[breaklines=true, breakindent=20pt]
# Nodes
Skill   # the procedure: purpose, trigger_when, scope_and_approval
Step    # a unit of work: name, description; an optional backing
        #   script; typed inputs/outputs; depends_on/parallel/one_of

# Edges between steps
inputs / outputs   # a step's typed output (name, type) feeds another
                   #   step's input -- the typed data-flow edges
depends_on         # explicit ordering for DAG-shaped procedures

# A step also binds package files
script             # a deterministic body under scripts/
references         # prose-cited docs under references/
\end{lstlisting}
\end{minipage}
\vspace{\baselineskip}

Steps may also carry \texttt{parallel} and \texttt{one\_of} control modifiers. The remaining top-level metadata---trigger and do-not-use conditions, anti-patterns, scenarios, modes, and integrations---attach as typed satellite nodes, keeping the entire specification queryable (Section~\ref{subsec:governance}).

\textbf{Compilation.} AIP includes a compiler meta-skill that transforms human-written source material such as existing skills, prose instructions, documentation, code, or informal descriptions into the graph representation. The compilation step acts as a quality gate where schema validation catches type errors, missing fields, and structural inconsistencies at authoring time rather than at runtime. Ambiguities in the source material must be resolved to produce a valid graph, which forces clarity that prose allows to remain implicit. The meta-skill also provides instruction on creating scripts from natural language where appropriate, which we have found can be a load-bearing step for improving reward. Some of the largest gains in our evaluation (Section~\ref{sec:eval}) are delivered by prose-heavy skills where the solver would otherwise re-derive code at run time, such as \texttt{mars-clouds-clustering} and \texttt{dapt-intrusion-detection}. Figure~\ref{fig:aipcompile} shows this transformation on a real skill, contrasting the human-curated and compiled on-disk packages.

\textbf{Execution.} At runtime, the agent loads the \texttt{SKILL.md} graph into its context window where it can reason through the execution graph steps. With more scripts and explicit input/output between steps, the agent can re-focus language-model reasoning on the prose nodes that genuinely require judgment and avoid re-deriving well-understood procedures on every run. The same principle of offloading deterministic computation to code is well-known to improve reliability~\cite{gao2023pal,chen2023pot}.

\textbf{Addressability.} Because each node is individually named, typed, and validatable, the graph is addressable at the component level. A failure in execution can be attributed to a specific node, the corresponding script can be inspected or corrected, and the repair can be validated in isolation. This property underlies both the skill-improvement loop evaluated in Section~\ref{sec:eval} and the governance, learning, and protocol roadmap outlined in Section~\ref{sec:discussion}.

\begin{figure*}[t]
\centering
\begin{minipage}[t]{0.40\textwidth}
{\footnotesize\textbf{Before: human-authored prose (\texttt{SKILL.md})}}\\[2pt]
\begin{lstlisting}
# Exoplanet Detection Workflows
## Pipeline Design Principles
### Key Stages
1. Data Loading: format, columns, time system
2. Quality Control: filter via quality flags
3. Preprocessing: remove noise, preserve signal
4. Period Search: choose algorithm for signal
5. Validation: verify candidate is real
6. Refinement: improve period precision

### Critical Decisions
Which period search algorithm?
- TLS: best for transit-shaped dips (box-like)
- Lomb-Scargle: any periodic signal, fast
- BLS: alternative to TLS, in Astropy
\end{lstlisting}
\end{minipage}\hfill
\begin{minipage}[t]{0.575\textwidth}
{\footnotesize\textbf{After: compiled AIP procedure (\texttt{SKILL.md}, YAML)}}\\[2pt]
\begin{lstlisting}
purpose: >
  Load, quality-filter, preprocess, search for a transit
  signal, validate vs SDE/SNR thresholds, refine period.
steps:
  - name: load-and-inspect
    outputs:
      - {name: lc-path,         type: string}
      - {name: column-map,      type: object}
      - {name: flag-good-value, type: float}
  - name: choose-search-strategy   # judgment node
    description: >
      Default to TLS; load references/method-selection.md
      to justify TLS vs Lomb-Scargle vs BLS.
    one_of:  [tls, lomb-scargle, bls]
    outputs: [{name: method, type: string}]
  - name: run-detection-pipeline   # script node
    script:  scripts/detect_period.py
    inputs:  [lc-path, column-map, flag-good-value]
    outputs: [{name: detection, type: object}]
  - name: validate-candidate       # script node
    script:  scripts/validate_candidate.py
    inputs:  [detection]
    outputs: [{name: verdict, type: object}]
\end{lstlisting}
\end{minipage}

\vspace{3pt}
{\footnotesize\textbf{On-disk skill package (before $\rightarrow$ after)}}\\[2pt]
\begin{minipage}[t]{0.40\textwidth}
\begin{lstlisting}
# Before: human-curated (prose-only)
exoplanet-workflows/
  SKILL.md     # prose guidance only

\end{lstlisting}
\end{minipage}\hfill
\begin{minipage}[t]{0.575\textwidth}
\begin{lstlisting}
# After: compiled AIP (exoplanet-workflows)
exoplanet-workflows/
  SKILL.md            # AIP procedure (YAML above)
  scripts/            # deterministic step bodies
    detect_period.py
    period_range_guide.py
    validate_candidate.py
  references/         # prose for judgment steps
    method-selection.md
    troubleshooting.md
  source/             # provenance kept by compiler
    original-SKILL.md
    procedure.schema.json
\end{lstlisting}
\end{minipage}
\caption{Compiling a prose skill to AIP, for \texttt{exoplanet-workflows}. The free-form procedure (top left) becomes a schema-validated YAML graph of typed steps (top right): each step declares typed \texttt{inputs}/\texttt{outputs}, binds to a deterministic script via \texttt{script}, or cites a prose \texttt{reference} from its description for judgment. On disk (bottom), the prose-only human skill---\texttt{exoplanet-workflows}, one of five modules the task bundles, all shipping no code---expands into a package whose steps are backed by generated \texttt{scripts/} and \texttt{references/}, with the original prose and JSON schema preserved under \texttt{source/}; Figure~\ref{fig:aipgraph} is this same skill projected as a graph in Neo4j.}
\label{fig:aipcompile}
\end{figure*}

\section{Evaluation}
\label{sec:eval}

\subsection{Experimental Setup}

\textbf{Benchmark.} We build on SkillsBench, a containerized agent benchmark of 94 tasks across 8 domains, run through the BenchFlow SDK~\cite{benchflow}. Each task bundles a natural-language instruction, a sandboxed environment, and a programmatic verifier that scores the agent's output. SkillsBench is designed to measure how an agent's task success changes when it is given reusable skills, and ships three native conditions per task: 1) \emph{noskill}, 2) one or more \emph{human-curated} skills authored offline by a domain expert, and 3) \emph{self-generated} one or more skills the agent writes for itself at trial time.

We extend SkillsBench with a multi-mode harness that adds two AIP conditions: 1) \emph{aip-from-instruction} where one or more AIP skills are authored by a Claude Code Agent from the task instruction alone, and 2) \emph{aip-from-curated} where one or more AIP skills are authored by a Claude Code Agent using the human-curated skill(s) as input. Both modes use the AIP meta-skill to compile the skills and both use Opus 4.7 as the language model.\footnote{The AIP protocol---its specification, schemas, and compiler meta-skill---is open source on GitHub at \url{https://github.com/zach-blumenfeld/aip} (the results in this paper use tag \texttt{v0.3a3}). The AIP-SkillBench harness, run data, and analysis are available on GitHub at \url{https://github.com/zach-blumenfeld/aip-skillbench}.} Following AIP's author-once, consume-many design, each AIP skill is authored a single time, committed, and mounted unchanged across all trials. The harness thus enables controlled head-to-head comparison of AIP skill \emph{formats} and \emph{authoring methods} under a single solver.

\textbf{Solver.} The agent harness is \texttt{claude-agent-acp}, driving the \texttt{claude-sonnet-4-6} model, with all executions isolated in a Docker sandbox and five independent trials per cell (one task under one condition).

\textbf{Conditions.} The primary comparison is between the \emph{human-curated} and \emph{aip-from-curated} modes. The harness modes (\emph{noskill} and \emph{selfgen}) are out of scope of the main claim. SkillsBench already found \emph{human-curated} outperforms both these modes on average. \emph{aip-from-instruction} was attempted in early trials, but results were excluded from this report. In the v0.3a2 medium set, \emph{aip-from-instruction} performed substantially lower than the human baseline (0/15 passing, vs. 11/15 for human-curated), with failures traced to incorrect methods in scripts committed into the authored skill---a distorting map projection, a parser that rejected valid inputs, a controller that produced no output. This reiterates SkillsBench's finding that self-authored skills provide no benefit on average---models cannot reliably author the procedural knowledge they benefit from consuming~\cite{skillsbench}---and that effective skills rest on expert curation~\cite{bakal2026knowledge}. It also leans toward the AIP format, rather than the authoring model (Claude Opus 4.7) being the source of the uplift in \emph{aip-from-curated}, though this does not fully isolate the two (see Limitations~\ref{sec:limitations}).

\textbf{Task sample and strata.} We evaluate on 27 tasks, listed in Table~\ref{tab:tasks}. We characterize each along three axes:
\begin{itemize}
  \item \textbf{Difficulty} (\emph{easy}, \emph{medium}, or \emph{hard}): the task author's rating, taken directly from SkillsBench task metadata.
  \item \textbf{Implementation class} (\emph{light} or \emph{heavy}): a task is \emph{heavy} when its declared type includes implementation, simulation, optimization, or control, i.e.\ it requires substantial code to be written or run; it is \emph{light} otherwise (analysis, calculation, extraction, search, or detection).
  \item \textbf{Structure class} (\emph{prose-only}, \emph{mixed}, or \emph{script-heavy}): a property of the \emph{human-curated} skill: how much runnable code it already ships. A \emph{prose-only} skill has no scripts (all natural language); a \emph{mixed} skill has scripts but more prose than code (measured in LoC); a \emph{script-heavy} skill has at least as much script as prose. This axis indexes how much room an AIP conversion has to add executability: most for prose-only skills, least for script-heavy ones.
\end{itemize}
The 24-task core is stratified across structure class $\times$ implementation class so the evaluation spans the gradient of expected AIP benefit rather than a single operating point, with difficulty and domain as secondary spread; all 24 use AIP v0.3a3. To this we add a three-task medium reference set (\emph{eval-3med}, marked $\dagger$ in Table~\ref{tab:tasks}). We report the full 27-task sample as the headline result, with the caveat that those three were compiled against an earlier protocol version (v0.3a2) and randomly selected (from the pool of medium difficulty tasks) rather than stratified. They were an earlier trial of experimentation. Results restricted to the 24-task v0.3a3 set are consistent in direction and significance (Section~\ref{sec:limitations}).

\begin{table*}[t]
  \centering
  \caption{The 27 evaluation tasks, grouped by the structure class of their human-curated skill (the axis indexing AIP's room to add executability). \emph{Diff.}\ is the SkillsBench difficulty rating; \emph{Impl.}\ is the implementation class (\emph{heavy} = the task type includes implementation, simulation, optimization, or control); \emph{Skills} is the number of curated skill modules the task bundles. The 24-task core is stratified across structure $\times$ implementation class; $\dagger$ marks the three supplementary \emph{eval-3med} tasks (v0.3a2 spec, hand-picked).}
  \label{tab:tasks}
  \footnotesize
  \begin{tabular}{@{}llcccp{5.9cm}@{}}
    \toprule
    Task & Domain & Diff. & Impl. & Skills & Description \\
    \midrule
    \multicolumn{6}{@{}l}{\textit{Prose-only --- human skill ships no scripts (12 tasks; most AIP room)}} \\
    \texttt{energy-unit-commitment}       & Industrial/Physical Systems & hard & heavy & 3 & Schedule generator unit commit for day-ahead demand. \\
    \texttt{mars-clouds-clustering}       & Natural Science & hard & heavy & 3 & Optimize unsupervised clustering of Mars cloud observations. \\
    \texttt{adaptive-cruise-control}      & Industrial/Physical Systems & med. & heavy & 5 & Implement and simulate an adaptive-cruise-control law. \\
    \texttt{bike-rebalance}               & Mathematics/Formal Reasoning & med. & heavy & 4 & Plan the optimal overnight redistribution of shared bikes. \\
    \texttt{drone-planning-control}$\dagger$ & Industrial/Physical Systems & med. & heavy & 6 & Generate drone trajectories and feedback control in simulation. \\
    \texttt{parallel-tfidf-search}        & Software Engineering & med. & heavy & 3 & Implement a parallelized TF-IDF document search. \\
    \texttt{fix-build-google-auto}        & Software Engineering & easy & light & 3 & Repair build errors in a Java codebase so it compiles. \\
    \texttt{offer-letter-generator}       & Office/White-Collar & easy & light & 1 & Fill a \texttt{.docx} offer-letter template with a conditional block. \\
    \texttt{enterprise-information-search} & Office/White-Collar & hard & light & 1 & Answer a retrieval query over enterprise documents. \\
    \texttt{spring-boot-jakarta-migration} & Software Engineering & hard & light & 5 & Migrate a Spring Boot codebase to Jakarta EE namespaces. \\
    \texttt{earthquake-plate-calculation}$\dagger$ & Natural Science & med. & light & 1 & Find the in-plate quake farthest from the Pacific boundary. \\
    \texttt{exoplanet-detection-period}   & Natural Science & med. & light & 5 & Detect an exoplanet and compute its orbital period. \\
    \midrule
    \multicolumn{6}{@{}l}{\textit{Mixed --- some scripts, but prose dominates (9 tasks)}} \\
    \texttt{energy-market-pricing}        & Industrial/Physical Systems & hard & heavy & 4 & Clear an energy market and compute locational prices. \\
    \texttt{grid-dispatch-operator}       & Industrial/Physical Systems & med. & heavy & 3 & Compute an economic unit dispatch for a grid. \\
    \texttt{jax-computing-basics}         & Software Engineering & med. & heavy & 1 & Implement numerical routines in JAX. \\
    \texttt{suricata-custom-exfil}        & Cybersecurity & med. & heavy & 3 & Author a Suricata rule for a custom exfiltration pattern. \\
    \texttt{court-form-filling}           & Office/White-Collar & easy & light & 1 & Extract case data and fill a court form. \\
    \texttt{powerlifting-coef-calc}       & Office/White-Collar & easy & light & 3 & Compute powerlifting scoring coefficients. \\
    \texttt{dapt-intrusion-detection}     & Cybersecurity & hard & light & 2 & Detect advanced-persistent-threat intrusion in PCAP traffic. \\
    \texttt{crystallographic-wyckoff-pos.}$\dagger$ & Natural Science & med. & light & 2 & Wyckoff position analysis from X-ray CIF files. \\
    \texttt{protein-expression-analysis}  & Natural Science & med. & light & 1 & Analyze cancer cell-line protein-expression data. \\
    \midrule
    \multicolumn{6}{@{}l}{\textit{Script-heavy --- skill is already executable and terse (6 tasks; least AIP room)}} \\
    \texttt{dialogue-parser}              & Software Engineering & easy & heavy & 1 & Parse dialogue text into a structured format. \\
    \texttt{civ6-adjacency-optimizer}     & Mathematics/Formal Reasoning & hard & heavy & 4 & Optimize district adjacency placement on a Civ\,VI map. \\
    \texttt{data-to-d3}                   & Software Engineering & med. & heavy & 1 & Build a D3.js (v6) visualization of stock data. \\
    \texttt{3d-scan-calc}                 & Industrial/Physical Systems & hard & light & 1 & Calculate the mass of a 3D-printed part from its geometry. \\
    \texttt{sec-financial-report}         & Finance/Economics & hard & light & 2 & Search SEC filings and analyze a financial report. \\
    \texttt{travel-planning}              & Mathematics/Formal Reasoning & med. & light & 6 & Plan an itinerary under scheduling constraints. \\
    \bottomrule
  \end{tabular}
\end{table*}

\textbf{Metrics.} The primary metric is mean task reward, which is robust to the ceiling and floor effects introduced by all-or-nothing verifiers. Secondary metrics are pass rate, wall-clock execution time, and tool-call count.

\subsection{Experimental Results}

\begin{figure*}[p]
  \centering
  \includegraphics[width=\textwidth]{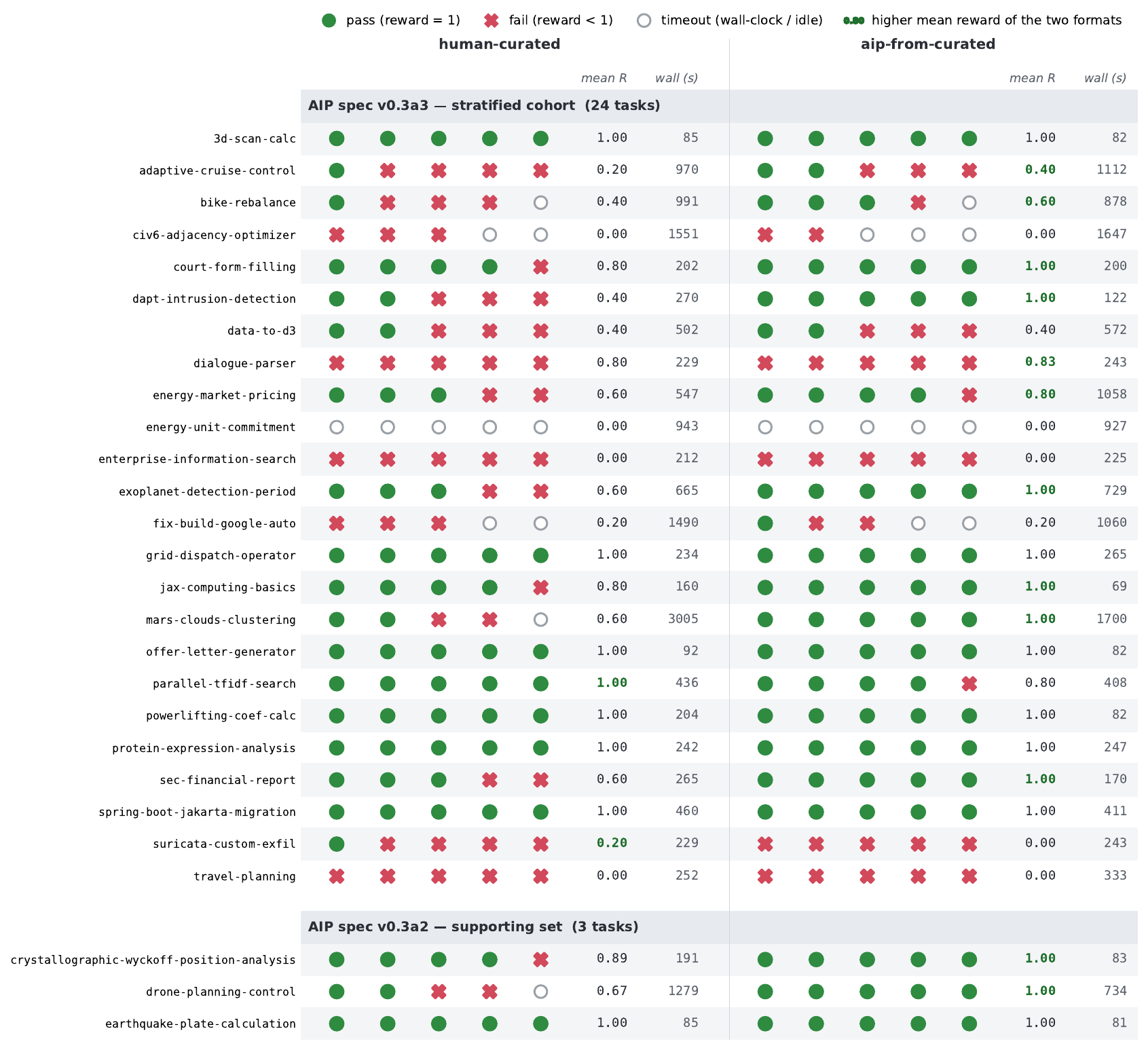}
  \caption{Per-trial outcomes for all 27 tasks under the two skill formats (five trials per task; within each cell the markers are sorted pass, fail, timeout), with the per-task mean reward and mean wall-clock (seconds) over those five trials shown to the right of each block. Each row is a task; the left block is the human-curated skill and the right block its AIP-compiled counterpart. A pass denotes a passing verifier (reward $=1$); all timeouts are trials terminated at the harness wall-clock cap or idle limit and score as reward~$0$, so the means are taken over all five trials. The higher mean reward of the two formats is shown in bold. Tasks are grouped by the AIP spec version their packs were compiled against and sorted alphabetically within each group: the 24-task stratified cohort (v0.3a3) and the three-task supplementary reference set (v0.3a2). Differentiating tasks (e.g.\ \texttt{dapt-intrusion-detection}, \texttt{mars-clouds-clustering}, \texttt{exoplanet-detection-period}) show failing or timed-out human-curated trials converted to passes after compilation; tied tasks are mutual ceilings or floors that no packaging could move.}
  \label{fig:trialmatrix}
\end{figure*}

Table~\ref{tab:main} reports the aggregate comparison between human-curated skills and their AIP-compiled counterparts. Compiling to AIP raises mean task reward from 0.599 to 0.705 ($+0.106$), a statistically significant gain under a Wilcoxon signed-rank test (\(p = 0.011\)), winning 12 tasks against 2 losses with 13 ties. Pass rate rises in parallel, from 53.3\% to 67.4\%. The 24-task v0.3a3 subset, which excludes the three hand-picked supplementary tasks, is consistent in direction and significance ($+0.101$, \(p = 0.022\)), confirming the headline does not hinge on the supplementary set. Figure~\ref{fig:trialmatrix} shows the trial-level outcomes behind these aggregates: the gains are concentrated in a subset of differentiating tasks, where compiling to AIP converts failing or timed-out trials into passes, while the many tied tasks reflect mutual ceilings (both formats pass all five trials) or mutual floors (both fail).

\begin{table}[t]
  \centering
  \caption{Human-curated vs.\ AIP-compiled skills on SkillsBench (Claude Sonnet solver, 5 trials/task). Reward is the primary metric; \(\Delta\) is AIP\,$-$\,human. The Wilcoxon signed-rank test is computed on per-task mean reward (\texttt{scipy}~\cite{virtanen2020scipy} default, tied pairs dropped). Wall-clock and tool-call deltas are descriptive: their aggregate differences are not statistically significant.}
  \label{tab:main}
  \small
  \begin{tabular}{lrrr}
    \toprule
    Metric & Human & AIP & $\Delta$ \\
    \midrule
    \multicolumn{4}{l}{\textit{27-task headline sample}} \\
    Mean task reward          & 0.599 & 0.705 & $+0.106$ \\
    Pass rate                 & 53.3\% & 67.4\% & $+14.1$\,pp \\
    Mean wall-clock (s)       & 585 & 510 & $-75$\textsuperscript{\dag} \\
    Mean tool calls           & 27.7 & 25.6 & $-2.1$\textsuperscript{\dag} \\
    Win / tie / loss          & \multicolumn{3}{c}{12 / 13 / 2} \\
    Wilcoxon $p$ (reward)     & \multicolumn{3}{c}{$\mathbf{0.011}$} \\
    \midrule
    \multicolumn{4}{l}{\textit{24-task v0.3a3 subset (robustness)}} \\
    Mean task reward          & 0.567 & 0.668 & $+0.101$ \\
    Pass rate                 & 50.8\% & 63.3\% & $+12.5$\,pp \\
    Win / tie / loss          & \multicolumn{3}{c}{10 / 12 / 2} \\
    Wilcoxon $p$ (reward)     & \multicolumn{3}{c}{$\mathbf{0.022}$} \\
    \bottomrule
  \end{tabular}
  \\[2pt]
  \footnotesize\raggedright
  \textsuperscript{\dag}Not statistically significant (per-task Wilcoxon \(p \approx 0.28\)); the mean reduction is driven by a few tasks. Pass rate is the fraction of trials with a passing verifier; mean reward is preferred because several verifiers are all-or-nothing, producing the high tie count.
\end{table}

\begin{figure}[t]
  \centering
  \includegraphics[width=\columnwidth]{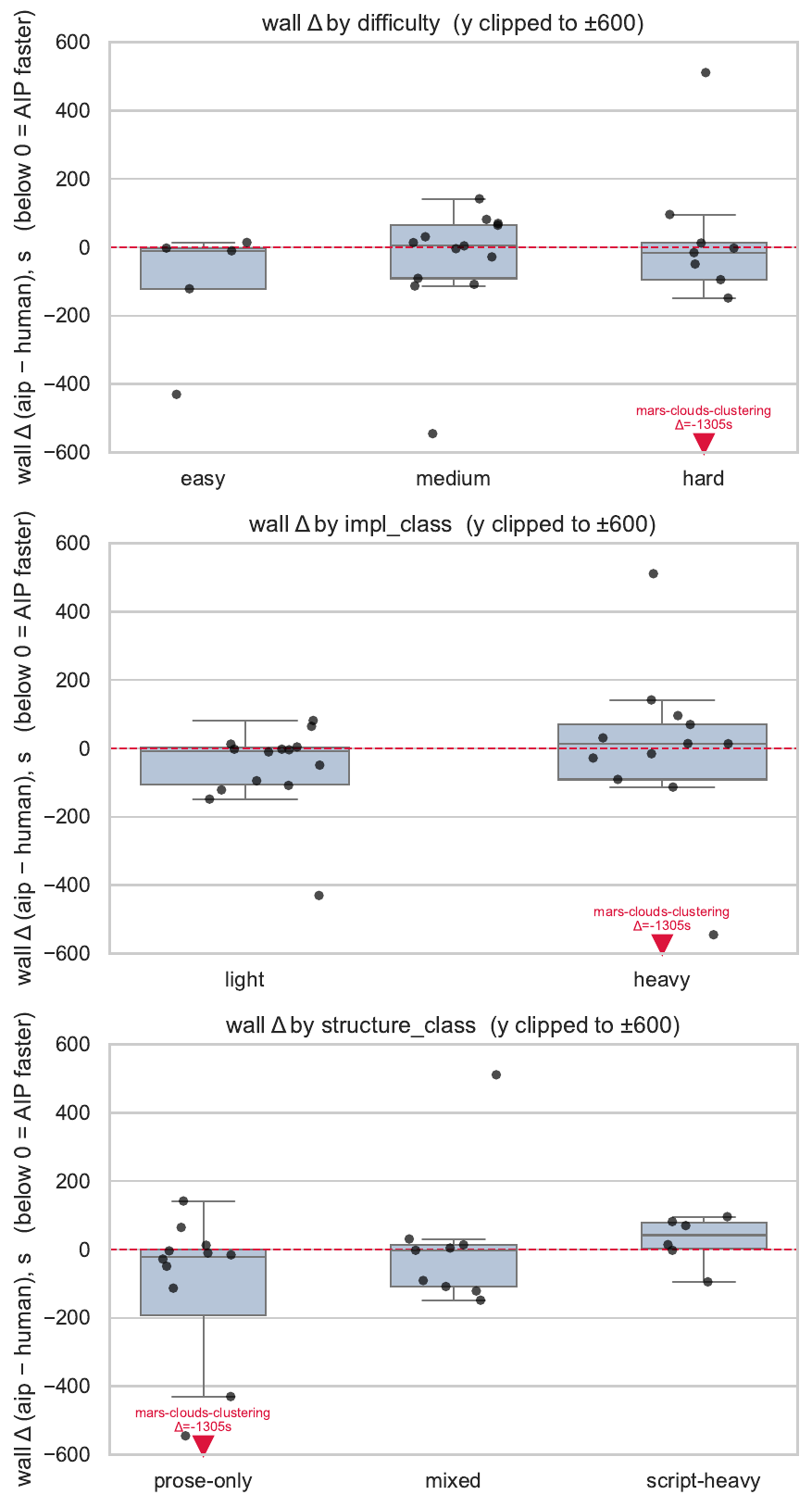}
  \caption{Per-task wall-clock change from compiling to AIP (\emph{aip} minus \emph{human} mean over five trials, in seconds) within each stratum; below the dashed line means AIP is faster. Boxes show the interquartile range and median; dots are individual tasks. The $y$-axis is clipped to $\pm 600$\,s, so \texttt{mars-clouds-clustering} ($-1305$\,s) is off-scale and flagged with a marker in the \texttt{hard}, \texttt{heavy}, and \texttt{prose-only} panels.}
  \label{fig:walldelta}
\end{figure}

\subsection{Execution time}
\label{sec:walltime}

AIP skills also run faster on average. Mean wall-clock time falls from 585\,s to 510\,s, and AIP is the faster format on 16 of 27 tasks---but this aggregate reduction is \emph{not} statistically significant (Wilcoxon \(p \approx 0.28\) two-sided, and \(p = 0.14\) under the directional hypothesis that AIP is faster). We therefore treat wall-clock and tool-call counts as descriptive, not as significance claims. The speedups are concentrated on a minority of tasks where prose forced the solver to re-derive code at run time (e.g.\ \texttt{dapt-intrusion-detection} 2/5\,$\to$\,5/5 at $\sim$2.2$\times$ faster, \texttt{jax-computing-basics} $\sim$2.3$\times$ faster).

Figure~\ref{fig:walldelta} breaks the per-task wall-clock change down by stratum. The clearest pattern is along \texttt{structure\_class}: prose-only skills, where AIP has room to add executable structure, tend to speed up, whereas already-terse script-heavy skills do not---mirroring the reward analysis, in which no single structural axis reaches significance.

Four tasks sit far outside the bulk of the distribution and are individually instructive; together they show that a wall-clock change is meaningful only when read alongside the reward outcome.

\textbf{\texttt{mars-clouds-clustering} (1305\,secs faster).} The human skill is prose-only (309 lines of reference text, no scripts) for a task that requires an 847-combination grid search over a clustering pipeline, scored by an all-or-nothing verifier. The human-curated solver re-derives the full pipeline on every trial and passes only 2 of 5: two completed runs produce a wrong result on an exact-specification detail, and a third computes the correct answer but is killed after idling at the harness time limit. The AIP conversion ships the vetted procedure and passes 5 of 5 while running roughly 30\% faster at equal load. The headline 1305\,secs slightly overstates the speedup, as two AIP trials ran under lighter concurrency.

\textbf{\texttt{drone-planning-control} (546\,secs faster).} Also prose-only (463 lines, no scripts). The human solver re-derives the trajectory-and-control stack from prose and passes 2 of 5, with one run stalling at the idle limit and two earning only partial credit; the AIP version passes 5 of 5 with markedly lower and tighter wall-clock.

\textbf{\texttt{fix-build-google-auto} (finished 430\,secs faster, by doing less work).} A build-repair task that both formats essentially fail (mean reward $0.20$ each): both arms thrash through 90--140 tool calls and exhaust the execution budget on multiple trials. AIP's lower mean wall-clock is an artifact of its unsuccessful trials terminating earlier, rather than a genuine efficiency gain---here the skill is not the bottleneck.

\textbf{\texttt{energy-market-pricing} (511\,secs slower but more accurate).} The lone task where AIP is markedly slower. The conversion added a heavier computational path that lands more correct results (4 of 5 vs.\ 3 of 5) but at roughly twice the tool calls and wall-clock. AIP's executability lever buys reliability at a compute cost that is not uniform across tasks.

\subsection{Skill improvement}
\label{sec:improvement}

Our results demonstrate that AIP skills are more \emph{executable}. They are also more \emph{improvable}, and for the same reason: because an AIP skill is a graph of named, typed, schema-validated nodes---each backed by a script that can be run and tested in isolation---a failure can be localized to a specific node and repaired easily, rather than by rewriting prose and hoping. We observed this loop directly while iterating the protocol from v0.3a2 to v0.3a3. Two skills that the compiler had authored with latent defects were diagnosed at the script level by an agent (Claude Code), corrected by a change to the AIP meta-skill, recompiled, and re-evaluated.

\textbf{\texttt{offer-letter-generator}: 0/5 $\to$ 5/5.} The compiled skill contained a frozen conditional-key bug, a template lookup keyed on \texttt{RELOCATION} rather than \texttt{RELOCATION\_PACKAGE}, so every trial failed in the same way. The defect was localized to a single node and fixed by a specification change (a functional-test correctness check, plus a key-suffix fallback and a default-keep rule) in this case. After recompilation the skill passed all five trials.

\textbf{\texttt{bike-rebalance}: 0.40 $\to$ 0.60, timeouts 3 $\to$ 1.} The compiler authored an over-engineered routing script (roughly 1{,}146 lines) heavy enough to exhaust the agent's time budget on three of five trials. A specification change favoring lean scripts produced a smaller routine that fit the budget, cutting timeouts and lifting reward.

Both repairs were verified to cause \emph{zero regressions} on the remaining tasks. The node-level edit fixed the target skill without disturbing the rest of the corpus. We executed these fixes by editing the AIP meta-skill and recompiling, but the loop is not bound to the meta-skill. The diagnosis was already agent-driven, and because each skill is a bounded, typed, addressable artifact---not just prose---the edit itself is a constrained, checkable action an agent can take directly on the skill, rather than the open-ended language editing where agents perform poorly. A team adopting AIP can therefore hand more skill maintenance to its own agents. Diagnosis, specification edit, recompilation, and re-evaluation thus form a closed feedback step with a measurable reward signal---which we argue in Section~\ref{subsec:rl} is the natural substrate for reinforcement learning over skills.

\subsection{Limitations}
\label{sec:limitations}

The evaluation, while yielding a statistically significant result, carries several important caveats that qualify the strength of our claims.

\textbf{Format--author confound.} The evaluation measures a compile-then-run pipeline: an agent compiles a human-written skill to AIP and then executes it. A performance gain could therefore reflect the improved graph representation, or it could reflect improvements made by agent scripting capability. The two mechanisms are not yet measured fully separately.

That said, the AIP spec---with typed inputs and outputs---shapes script design and appears to contribute to evaluation and improvement. The two node-level repairs of Section~\ref{sec:improvement} appear to be guided by the typed specification: because the spec isolated and sequenced steps as connected nodes, the agent could localize the brittleness to a specific script and address it through a functional-testing rule in the meta-skill. This suggests the typed representation, not merely scripting, carries part of the effect. A clean separation nonetheless requires the ablation described in Section~\ref{sec:control}.

\textbf{Statistical power.} Trials are limited to $n = 5$ per cell, and the Wilcoxon signed-rank test operates on per-task means over 12--14 non-tied pairs. The result is significant but preliminary and any broader claims should await a wider trial budget and a larger task set.

\textbf{Verifier characteristics.} Several tasks use all-or-nothing verifiers, which inflates per-task variance and produces the high number of ties (12--13). A finer-grained reward signal would provide more discriminative power and reduce ceiling and floor effects.

\textbf{Subset caveats.} Three of the 27 headline tasks (the \emph{eval-3med} reference set) were compiled against an earlier protocol version (v0.3a2 rather than v0.3a3) and were randomly sampled from medium difficulty tasks rather than drawn from the stratified sampling procedure. We therefore report both the full 27-task figure and the 24-task v0.3a3 subset, and the two are consistent in direction and significance: $+0.106$ mean reward at $p = 0.011$ over 27 tasks, versus $+0.101$ at $p = 0.022$ over the 24-task stratified set. Accordingly, the headline result does not depend on the supplementary tasks.

\textbf{Budget-capped tasks.} Two tasks (energy unit commitment and Civilization~6) hit the execution budget without producing informative results. They are excluded in spirit, though they land as ties in the aggregate counts. These should either be re-budgeted or dropped in future iterations.

\textbf{Single model.} All experiments use Claude Sonnet as the solver. Two replications are needed. First, a weaker model (e.g., Claude Haiku) would test the hypothesis that graph structure provides proportionally greater benefit to less capable models. Second, models from other vendors such as OpenAI's GPT or Google's Gemini, and open-weight families such as Llama, Qwen, DeepSeek, and Mistral would help to establish that the gains hold across LLMs rather than being specific to one model family.

\section{Related Work}

\textbf{Agent skills and their representations.}
ReAct~\cite{yao2023react} interleaves language-model reasoning with tool actions; Voyager~\cite{wang2023voyager} accumulates learned behaviors as code snippets indexed by prose; and Toolformer~\cite{schick2023toolformer} teaches models to invoke external APIs mid-generation. Anthropic's Agent Skills~\cite{anthropic-agent-skills,agentskills-spec} standardize the packaging of such procedural knowledge, which a growing literature treats as institutional or expert knowledge to be transferred to agents~\cite{bakal2026knowledge} and surveys along axes of architecture and acquisition~\cite{xu2026agentskills}. Closest to our work, SSL~\cite{liang2026ssl} also argues for structuring skill artifacts, but targets skill \emph{discovery and assessment} rather than execution. These package procedural knowledge differently---as prose (Agent Skills), as code retrieved by prose descriptions (Voyager), or as structure aimed at skill discovery (SSL)---but none gives the skill a typed, schema-validated graph of scripted and prose steps, nor measures its effect on task execution; that is what AIP contributes.

\textbf{Structured execution and workflow graphs.}
PAL~\cite{gao2023pal} and Program of Thoughts~\cite{chen2023pot} offload deterministic computation to code while reserving language-model reasoning for the rest, and chain-of-thought prompting~\cite{wei2022cot} externalizes reasoning as explicit intermediate steps.

At the system level, GPTSwarm~\cite{zhuge2024gptswarm} casts language agents as optimizable computational graphs whose node prompts and edge connectivity are refined automatically by search or reinforcement learning---an early example of the agentic-computation-graph paradigm surveyed by Yue et al.~\cite{yue2026acgsurvey}, where the graph is the orchestration of a specific agent system, machine-optimized for that system. AIP instead places the graph at the level of a portable skill \emph{artifact}, compiled from a human-authored skill and consumed by any general-purpose agent rather than bound to one bespoke system.

In practice, agent frameworks such as LangGraph and Google's ADK~\cite{langgraph,google-adk} represent agent behavior as graphs of steps, DSPy~\cite{khattab2023dspy} compiles declarative pipelines instead of hand-tuning prompts, and predefined workflow structure is both recommended~\cite{anthropic-effective-agents} and benchmarked~\cite{flowbench} as a route to reliability. AIP brings this graph view to the skill specification itself: scripted nodes carry deterministic work and prose nodes carry judgment, connected by typed input/output edges. These systems and AIP differ in kind, not degree (Table~\ref{tab:novelty}). LangGraph, ADK, and DSPy are runtime orchestration frameworks that an engineer programs in code to build a specific agent. AIP instead defines a portable skill \emph{artifact} that a domain expert authors in prose and a compiler translates, consumed by a general-purpose agent rather than executed by a bespoke runtime. Where those frameworks require authoring a graph from scratch, AIP compiles the large existing corpus of human-written prose skills into graph form, and---because it extends the Agent Skills specification---remains model- and vendor-agnostic with no runtime dependency. It also means that adding additional procedural knowledge comes at near-zero marginal cost, since AIP skills are auto-discovered and loaded on demand through progressive disclosure without any required changes to runtime orchestration code. DSPy is complementary but orthogonal in target: it optimizes the prompts or weights of a developer-defined pipeline, whereas AIP structures the skill representation itself.

This artifact-level, schema-validated typing is what enables the properties we rely on: node-addressable diagnosis and repair, with corpus-level governance queries and a bounded action space for learning over skills as future work.

\begin{table}[htbp]
  \centering
  \caption{AIP vs.\ existing graph/workflow systems. AIP operates at the level of a portable, schema-validated \emph{skill artifact} compiled from human prose, rather than engineer-written runtime orchestration.}
  \label{tab:novelty}
  \footnotesize
  \begin{tabular}{@{}lcccc@{}}
    \toprule
     & LangGraph/ & DSPy & Agent & AIP \\
     & ADK &  & Skills &  \\
    \midrule
    Authored by        & engineer & engineer & expert  & expert \\
                       & (code)   & (code)   & (prose) & +compiler \\
    Source material    & scratch  & scratch  & knowledge & existing skills \\
    Portable artifact  & no       & no       & yes     & yes \\
    Typed I/O edges    & yes      & partial  & no      & yes \\
    Schema-validated   & partial  & no       & no      & yes \\
    Corpus-queryable   & no       & no       & no      & yes \\
    \bottomrule
  \end{tabular}
\end{table}

\textbf{Benchmarking agentic systems.}
AgentBench~\cite{liu2023agentbench} and SWE-bench~\cite{jimenez2024swbench} evaluate agents on real-world tasks with programmatic verifiers. SkillsBench, run through the BenchFlow SDK~\cite{benchflow}, instead isolates the incremental effect of a reusable skill by scoring each task with no skill, a curated skill, and a self-generated skill. We extend SkillsBench with AIP conditions to compare skill \emph{formats} under a single solver.

\textbf{Editing, self-improvement, and learning over skills.}
Editing a skill is unreliable for the reasons set out earlier---the content is unfamiliar and free-form prose offers no bounded surface to edit against, so model edits skew additive and intrinsic self-revision without feedback rarely helps~\cite{sclar2024formatspread,singhal2024length,huang2024selfcorrect}. A parallel line pursues agents that improve themselves~\cite{gao2025selfevolving,zweiger2025seal,robeyns2025selfimproving}, building on reinforcement learning from a reward signal~\cite{ouyang2022instructgpt,sutton2018rl}. AIP's typed, schema-validated graph turns a skill edit into a bounded, checkable action and supplies the reward-bearing feedback loop these methods need at the level of an individual skill.

\textbf{Governing skill corpora.}
As agentic systems scale, auditability and accountability become first-order concerns~\cite{saini2026governing}. Since an AIP skill is a typed graph, a corpus of skills can be projected into a graph database~\cite{robinson2015graphdb} and queried for skills missing an approval step, shared sub-procedures, or reusable templates. This moves governance from manual documentation review to structured query and deterministic curation.

\section{Discussion and Future Work}
\label{sec:discussion}

\subsection{Experimental control}
\label{sec:control}
The highest-value missing experiment is an ablation separating the graph
representation from agent scripting, in two arms. First, a \emph{flattened}
arm holds scripts fixed: each AIP-compiled package retains its
\texttt{scripts/} and \texttt{references/} unchanged, with only the
\texttt{SKILL.md} rewritten as plain-Markdown prose, so any delta against
AIP measures the runtime value of the typed graph alone. Second, a
\emph{script-conversion} arm compiles scripts without the AIP spec,
delivering the \texttt{SKILL.md} as a plain-Markdown Agent Skill rather
than a graph. Because the AIP spec---with typed inputs and outputs---shapes
script design, comparing this arm against the first measures the spec's
authoring-time contribution. Together these disentangle graph representation
from agent scripting alone. Our immediate future work will be to establish these
arms as baselines, and confirm the promising behaviors observed in this
paper.

\subsection{From per-skill graphs to corpus governance}
\label{subsec:governance}
The same typed graph that makes a skill's deterministic steps runnable also makes the skill queryable. A library of AIP skills can be projected into a graph database, enabling audits such as identifying skills that lack an approval or validation step, discovering skills that share a common sub-procedure, and composing skills from reusable node templates. This moves skill governance from a manual documentation review to a structured query over a typed graph. An open question is whether providing an agent with query access to a skill corpus, rather than delivering a skill in-context as YAML, further improves task performance. We identify a controlled A/B experiment on this as a high-value next step.

\subsection{Reinforcement learning over the skill graph}
\label{subsec:rl}
Left unconstrained, autonomous skill writing tends to accrete prose and code without pressure toward compression or toward the right boundary between scripted and natural-language nodes. The AIP execution graph provides a bounded, typed, validity-gated action space: edits are changes to nodes, scripts, or edges, each of which can be validated against the schema and evaluated against a reward signal. The manual revision cycle from v0.3a2 to v0.3a3---in which failures were diagnosed at the node level, corrections were made to the specification, and the updated skill was re-evaluated---is a policy step in this framework. Automating the edit--evaluate--edit loop constitutes reinforcement learning over skills~\cite{sutton2018rl,ouyang2022instructgpt}. The node-level repair results of Section~\ref{sec:improvement} serve as proof-of-concept that the feedback signal is both localizable and actionable. Formalizing this loop is a key topic for future work.

\subsection{From specification to protocol}
\label{subsec:protocol}
Today, AIP is closer to a specification than a protocol: the agent loads the entire YAML graph into its context window and follows the traversal logic through its own reasoning, with nothing enforcing adherence to the graph topology. This keeps AIP compatible with current agent-skill formats and lets us benchmark it, but it leaves headroom. Since traversal is unenforced, reliable execution still rests on the agent's own discipline, which is a ceiling that an enforced protocol could raise, especially for smaller models. Loading the full specification into context on every run also adds a token burden that can raise cost and erode performance through context rot.
We quantify this cost in Table~\ref{tab:tokencost}: compiling to AIP increases the
mean specification a task loads by roughly 78\%, yet mean wall-clock \emph{falls}
from 585\,s to 510\,s (\S\ref{sec:walltime})---the added context does not make runs
slower in practice.\footnote{The clearest measure of cost in practice is the total
number of tokens a task consumes end to end---dominated by the agent's multi-turn
reasoning rather than the one-time specification. The SkillsBench harness does not
log per-call token usage, so we report the loaded specification size and wall-clock
here and leave full per-task token accounting to future work.}

Since AIP already defines a typed action surface, a full protocol for walking the graph is feasible. It would execute nodes through controlled local or remote calls rather than in-context reasoning, while remaining backward-compatible with the agent-skill specification. We consider this the most important consideration for future work.

\begin{table}[htbp]
  \centering
  \caption{Mean context-token and latency cost of AIP-compiled vs.\ human-curated
  skills, per task ($n{=}27$). SKILL.md is the specification loaded when a skill is
  invoked; references are loaded on demand. Compiling to AIP enlarges the loaded
  specification, yet end-to-end latency does not rise.}
  \label{tab:tokencost}
  \small
  \begin{tabular}{lrrr}
    \toprule
     & Human & AIP & $\Delta$ \\
    \midrule
    SKILL.md (tokens)   & 4{,}860 & 8{,}669 & $+3{,}809$ \\
    references (tokens) & 1{,}792 & 4{,}366 & $+2{,}574$ \\
    wall-clock (s)      & 585     & 510     & $-75$ \\
    \bottomrule
  \end{tabular}
\end{table}

\section{Conclusion}
\label{sec:conclusion}

Agent skills are written largely as free-form prose, resulting in the agent re-deriving procedures in every session that could have been captured with structure and code. This has a cost in terms of reliability and consistency on implementation-heavy tasks. Editing prose is something both humans and agents do poorly, for the distinct reasons set out in Section~\ref{sec:intro}. AIP compiles a human-written skill into a directed execution graph: scripted where computation is deterministic, natural language where judgment is needed, connected by typed input/output edges and schema-validated throughout. Experts still author the skill; AIP makes its deterministic steps runnable and every step addressable.

The result is a representation that is more executable and improvable. Compiling human-written skills to AIP yields a statistically significant gain in task reward on SkillsBench, by handing the agent vetted, runnable units and a fixed procedure rather than asking it to re-plan from prose. Because each skill is a graph of named, typed, testable nodes, a failure localizes to a single node and is repaired by a checkable edit---a loop we ran by hand here, but one an adopter's agent can run directly. The same structure makes a skill corpus queryable for governance and provides a bounded, typed action space for reinforcement learning over skills.

AIP today is a specification an agent reads and follows; enforcing that traversal as a runtime protocol is a potential next step. But the core lesson already holds: a skill that is a graph is easier to execute, easier to diagnose when it fails, and easier to improve once diagnosed. The graph is not incidental to these properties---it is their common cause, and the foundation for skill libraries that can be reliably executed, governed, and learned.


\bibliographystyle{ACM-Reference-Format}
\bibliography{references}

\end{document}